\documentclass[11pt,a4paper]{article}
\usepackage[hyperref]{acl2019}
\usepackage{times}  %Required
\usepackage{helvet}  %Required
\usepackage{courier}  %Required
\usepackage{url}  %Required
\usepackage{graphicx}  %Required
\usepackage{latexsym}
\usepackage{subcaption}
\usepackage{url}
\usepackage{times,latexsym,arydshln,amsmath,graphicx,xcolor,multirow}
\usepackage{amsfonts, amssymb}

\def\sclass{S-class}
\def\uniform{unif$\Sigma$}
\def\weighted{wght$\Sigma$}

\newcounter{notecounter}
\newcommand{\enotesoff}{\long\gdef\enote##1##2{}}
\newcommand{\enoteson}{\long\gdef\enote##1##2{{
\stepcounter{notecounter}
{\large\bf
\hspace{1cm}\arabic{notecounter} $<<<$ ##1: ##2
$>>>$\hspace{1cm}}}}}
\enoteson
\enotesoff

\def\figref#1{Figure~\ref{fig:#1}}
\def\figlabel#1{\label{fig:#1}\label{p:#1}}

\def\tabref#1{Table~\ref{tab:#1}}
\def\tablabel#1{\label{tab:#1}\label{p:#1}}

\def\secref#1{\S\ref{sec:#1}}
\def\seclabel#1{\label{sec:#1}}
\def\eqref#1{Eq.~\ref{eqn:#1}}

\newcommand*{\affaddr}[1]{#1} % No op here. Customize it for different styles.
\newcommand*{\affmark}[1][*]{\textsuperscript{#1}}

\usepackage{url}

\usepackage{color}
\usepackage{xcolor}

\aclfinalcopy % Uncomment this line for the final submission
\def\aclpaperid{1782} %  Enter the acl Paper ID here

\long\def\eat#1{\ignorespaces}

 \title{Probing for Semantic Classes:\\
   Diagnosing the Meaning Content of Word Embeddings}
\author{{\centering
Yadollah Yaghoobzadeh{\rm\affmark[1]} ~~
Katharina Kann{\rm\affmark[2]} ~~ 
Timothy J. Hazen{\rm\affmark[1]} ~~
Eneko Agirre{\rm\affmark[3]} ~~
Hinrich Sch\"{u}tze{\rm\affmark[4]}} \vspace{.15cm}
\\
\affaddr{\affmark[1]Microsoft Research Montr\'eal}\\
\affaddr{\affmark[2]Center for Data Science, New York University}\\
\affaddr{\affmark[3]IXA NLP Group, University of the Basque Country} \\
\affaddr{\affmark[4]CIS, LMU Munich} \\
\affaddr{\texttt{yayaghoo@microsoft.com}}}

\begin{document}
\maketitle

\begin{abstract}
Word embeddings typically represent different meanings of a word in a
single conflated vector. Empirical
analysis of embeddings of ambiguous words is currently
limited by the small size of manually annotated resources
and by the fact that word senses are treated as unrelated
individual concepts. We present a large
dataset based on manual Wikipedia annotations and word
senses, where word senses from different words are related
by semantic classes.
This is the basis for novel diagnostic tests for an embedding's
content: we
\emph{probe word embeddings for semantic classes} and
analyze the embedding space by classifying embeddings into semantic
classes. Our main findings are: (i) Information
about a sense is generally represented well in a single-vector
embedding -- if the sense is frequent. (ii) A classifier can
accurately predict whether a word is single-sense or multi-sense, based only on
its embedding.
(iii) Although rare senses are not well represented in
single-vector embeddings, this does not have negative impact
on an NLP application whose performance depends on frequent senses.
\end{abstract}

% e.g.,
% part-of-speech tagging \cite{tsvetkov2016pos}, text similarity \cite{kenter2015short}, 
% sentence classification \cite{kim2014conv} and knowledge base completion (KBC) \cite{wang2014knowledge}.

\section{Introduction}
Word embeddings learned by methods like Word2vec \cite{mikolov2013efficient}
and Glove \cite{glove14} have had a big impact on
natural language processing (NLP)
and information retrieval (IR). 
They are effective and efficient for many tasks.  
More
recently, contextualized embeddings like ELMo
\cite{elmo2018} and BERT \cite{bert} have further improved
performance. 
To understand both word and
contextualized embeddings, which still rely on word/subword embeddings at their lowest layer,  we must peek inside the blackbox embeddings.
% for word
% embeddings, including embeddings at the lowest layer of
% models like ELMo and BERT.

Given the importance of word embeddings, attempts have
been made to construct diagnostic tools to analyze them. 
However, the main tool for  analyzing their semantic
content is still
looking at  nearest neighbors of embeddings. Nearest
neighbors are based on full-space similarity neglecting the
multifacetedness property of words \cite{gladkova16} and
making them unstable \cite{instability2018}.

As an alternative,
\emph{we propose
  diagnostic classification of embeddings into semantic classes
  as a probing task to reveal their meaning content.}
We will refer
to semantic classes as \emph{\sclass es}.
We use \sclass es such as \texttt{food}, \texttt{drug} and \texttt{living-thing} to define word 
senses. \sclass es are frequently used for semantic
analysis, e.g., by
\citet{Kohomban:2005}, \citet{Ciaramita:2006} and
\citet{Izquierdo:2009} for word sense disambiguation, but
have not been used for analyzing
embeddings.

Analysis based on \sclass es is only promising if we have
high-quality \sclass\ annotations.  Existing datasets are
either too small to train  embeddings, e.g., SemCor
\cite{Miller1993semcor}, or artificially generated
\cite{derata16}.  Therefore, \emph{we build WIKI-PSE, a
  WIKIpedia-based resource for Probing Semantics in word
  Embeddings}.  We focus on common and proper
nouns, and use their \sclass es
as proxies for senses.  For example, ``lamb''
has the senses \texttt{food} and \texttt{living-thing}.
  
Embeddings do not explicitly address ambiguity;
multiple senses of a word are crammed into a single vector.
This is not a problem in some applications
\cite{li2015}; one possible explanation 
is that this is an effect of
sparse coding that supports the recovery of individual meanings from
a single vector \cite{arora18tacl}.
But ambiguity has an adverse effect in other scenarios,
e.g., \newcite{xiao14crosslingual} see the need of filtering out embeddings of ambiguous
words in dependency parsing.

% There is currently no good methodology for empirically
% analyzing ambiguity in word embeddings. 

We present the first  comprehensive empirical analysis of ambiguity in word embeddings.
Our resource, WIKI-PSE,  enables novel diagnostic tests that help
explain how (and how well) embeddings represent multiple meanings.\footnote{WIKI-PSE is 
available publicly at \url{https://github.com/yyaghoobzadeh/WIKI-PSE}.
}

%So we train diagnostic classifiers that
% predict from an embedding which semantic classes the word has. 
%Since  semantic classes are shared across words, we can learn
%  robust classifiers for them that establish with high accuracy what
%  information the embedding of an ambiguous word contains.

Our diagnostic tests
show: (i) Single-vector  embeddings can
represent many non-rare
senses well.
(ii) A classifier can
accurately predict whether a word is single-sense or multi-sense, based only on
its embedding.
(iii)
In experiments with  five
common datasets for mention, sentence and sentence-pair classification tasks,
the lack of representation of rare senses in single-vector
embeddings has little negative impact -- 
this indicates that for many common NLP benchmarks only
frequent senses are needed.

\section{Related Work}
\sclass es (semantic classes) are a central concept in semantics and in the analysis of
semantic phenomena \cite{yarowsky92,Ciaramita2003,semcat2018}. 
They have been used for analyzing
ambiguity
by 
\citet{Kohomban:2005}, \citet{Ciaramita:2006}, and
\citet{Izquierdo:2009}, \emph{inter alia}.
There are some datasets designed for interpreting word
embedding dimensions using \sclass es,
e.g., 
SEMCAT \cite{semcat2018} and HyperLex \cite{hyperlex}.
The main differentiator of our work is our probing approach using
supervised classification of word embeddings. 
Also, we do not use WordNet senses but 
Wikipedia entity annotations since WordNet-tagged corpora are small.

In this paper, we probe
word embeddings with supervised classification.
Probing the layers of neural networks has become very popular.
\newcite{conneau2018you} probe sentence embeddings on how well they predict
linguistically motivated classes. 
\newcite{hupkes2018visualisation}
apply diagnostic classifiers to test hypotheses about the
hidden states of RNNs.
Focusing on embeddings, \newcite{kann-etal-2019-verb} investigate how well sentence and word representations encode information necessary for inferring the idiosyncratic frame-selectional properties of verbs. Similar to our work, they employ supervised classification.
\newcite{tenney2018you} probe syntactic and semantic information learned 
by contextual embeddings  \cite{melamud2016context2vec,mccann2017learned,elmo2018,bert} compared to non-contextualized embeddings. 
They do not, however, address ambiguity, a key phenomenon of language. 
While the terms ``probing'' and ``diagnosing'' come from this
literature, similar probing experiments were used in earlier work, e.g., \newcite{derata16} probe for linguistic properties in word embeddings using synthetic data and
also the task of corpus-level fine-grained entity typing  \cite{figment15}.

%, applied to word embeddings. 
%, and we use diagnostic classifiers to analyze the
%effect of ambiguity on word embeddings.  
%As opposed to
%sentence probing tasks, it is more challenging to find good
%probing tasks for words.  Sentence embeddings are mostly
%compositions of their content, so probing tasks can be
%defined for the surface and syntactic forms of sentence
%constituents.  Words do not have such observable
%semantic constituents and so it is not easy to define
%semantic probing tasks
%for them.
%

We use our new resource WIKI-PSE for analyzing
ambiguity in the word embedding space.  Word sense
disambiguation (WSD)
\cite{Agirre:2007:WSD:1564561,navigli2009wsd} and entity linking (EL)
\cite{bagga1998algorithms,wikify2007} are related
to ambiguity in
that they predict
the context-dependent sense of an ambiguous word or entity.
In our complementary approach, we
analyze directly how multiple senses are represented in
embeddings.  While WSD and EL are important, they
conflate (a) the evaluation of the information content of an
embedding with (b) a model's ability to extract that
information based on contextual clues.  
We mostly focus on (a) here. 
Also, in contrast to WSD
datasets, WIKI-PSE  is not based on inferred
sense tags and not based on artificial ambiguity, i.e.,
pseudowords \cite{gale1992work,schutze92}, but on real
senses marked by Wikipedia hyperlinks.
There has been work in generating dictionary definitions from word embeddings
\cite{noraset2017definition,bosc2018auto,artyom2018}.
\newcite{artyom2018} explicitly adress ambiguity and generate definitions for words
conditioned on their embeddings and
selected contexts.
This also conflates (a) and (b).

Some prior work also looks at how ambiguity affects word embeddings.
\newcite{arora18tacl} posit that a word embedding is a linear
combination of its sense embeddings and that senses can be
extracted via sparse coding. 
\newcite{mu2016geometry} argue
that sense and word vectors are linearly related and
show that word embeddings are intersections of sense
subspaces.  Working with synthetic data,
\newcite{derata16} evaluate embedding models on how robustly
they represent two senses for low vs.\ high skewedness of senses.
Our analysis framework is novel and complementary,
with
several new findings. 

%Parametric sense embedding
%learning models have a fixed number of senses per word
%\cite{huang2012glove,tian2014}.  
%Non-parametric models
%induce the number of senses
%\cite{schutze98,neelakantan2014,li2015}. 
%See
%\newcite{sense-survey2018} for a survey.  

Some believe
that ambiguity should be eliminated from
embeddings, i.e., that a separate embedding is
needed for each sense \cite{schutze98,huang2012glove,neelakantan2014,li2015,sense-survey2018}. 
This can improve performance on
contextual word
similarity, but a recent
study \cite{dubossarsky2018emnlp} questions this finding.
WIKI-PSE allows us to compute sense embeddings; we will
analyze their effect on word embeddings in our diagnostic
classifications.
% Contextualized word embeddings \cite{melamud2016,mccann2017learned,elmo2018,bert} implicitly
% and seamlessly
% model sense embeddings as function of context words.
% The input layer of all these models is still embeddigns representing multiple
% meanings.

\begin{figure}
\centering
\includegraphics[width=.5\textwidth]{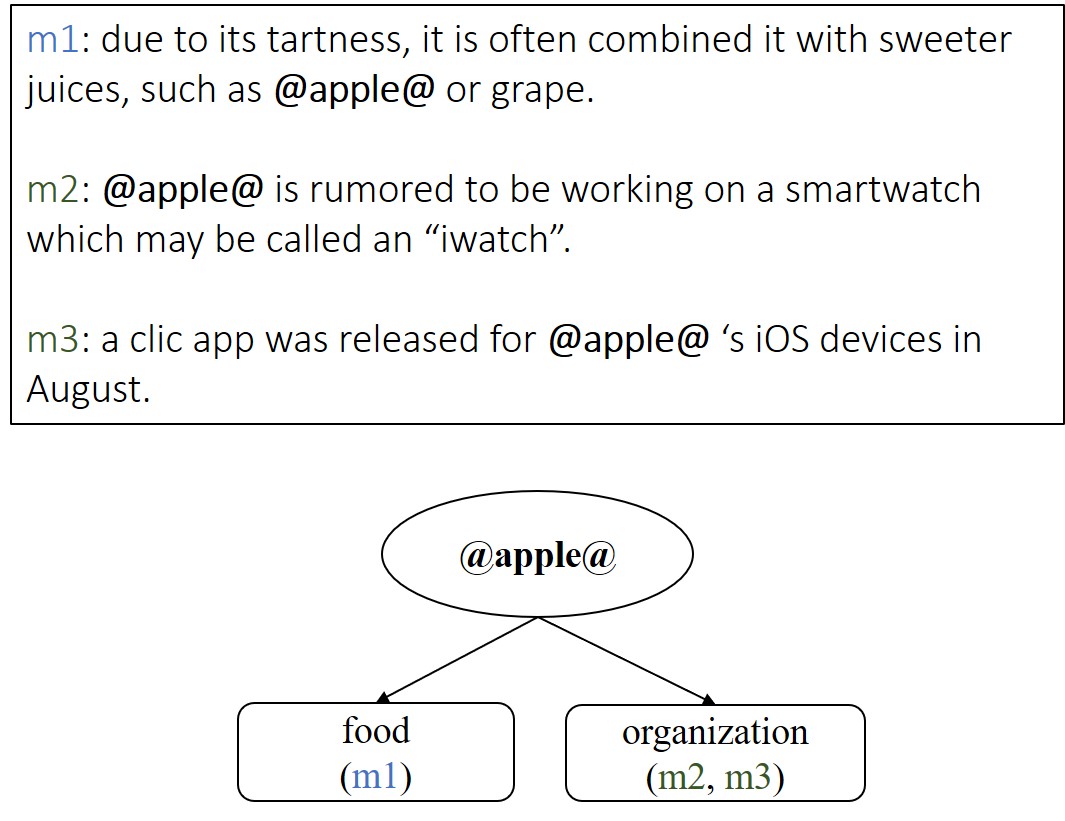}
\caption{Example of how we build WIKI-PSE. 
There are three sentences linking ``apple''
 to different entities. 
There are two mentions ($m_2$,$m_3$) with the
\texttt{organization} sense (\sclass) and one mention ($m_1$)
with the \texttt{food} sense (\sclass).
}
\figlabel{apple}
\end{figure}

\section{WIKI-PSE Resource}
\seclabel{dataset}
We want to create a resource that allows us to probe
embeddings for \sclass es. Specifically, we have the
following desiderata:\\
(i) We need a corpus that is \sclass-annotated at the token
level, so that we can train sense embeddings
as well as conventional word embeddings.
(ii) We need a dictionary of the corpus vocabulary that is
\sclass-annotated at the type
level. This gives us 
a gold standard for probing
  embeddings for \sclass es.\\
(iii)
 The resource must be large so that we have a
 training set of sufficient size that lets us
compare different embedding learners and 
 train
    complex models for probing.

 We now describe WIKI-PSE, a Wikipedia-driven resource for
 Probing  Semantics in Embeddings, that satisfies our desiderata.

WIKI-PSE consists of a corpus and a corpus-based dataset of word/\sclass\
pairs: an \sclass\ is assigned to a word if the
word occurs with that \sclass\ in the corpus.  
There exist sense annotated corpora like SemCor \cite{Miller1993semcor},
but due to the cost of annotation, those corpora are
usually limited in size, which can hurt the quality
of the trained word embeddings -- an
important factor for our analysis.

In this work, we propose a
novel and scalable approach to building a corpus without depending on manual
annotation except in the form of Wikipedia anchor links. 

WIKI-PSE  is based on the English Wikipedia (2014-07-07).
Wikipedia is suitable for our purposes since
it contains nouns -- proper and common nouns -- disambiguated and linked to 
Wikipedia pages via anchor links.  To find more abstract
meanings than Wikipedia pages, we annotate the 
nouns with \sclass es.  
We make use of the 113 FIGER types\footnote{We follow the mappings in \url{https://github.com/xiaoling/figer} to first find the corresponding Freebase topic of
a Wikipedia page and then map it to FIGER types.} \cite{ling2012fine}, e.g., \texttt{person} and \texttt{person/author}.

\begin{table}
\footnotesize
\centering{
\begin{tabular}{p{2.5in}l}
location, person, organization, art, event, broadcast\_program, 
title, product, living\_thing, people-ethnicity, language, 
broadcast\_network, time, religion-religion, award, 
internet-website, god, education-educational\_degree, food, 
computer-programming\_language, metropolitan\_transit-transit\_line, transit, finance-currency, disease, chemistry, body\_part, 
finance-stock\_exchange, law, medicine-medical\_treatment, 
medicine-drug, broadcast-tv\_channel, medicine-symptom, biology, 
visual\_art-color 
\end{tabular}
}
\caption{\sclass es in WIKI-PSE sorted by frequency.}
\tablabel{types}
\end{table}

Since we use distant supervision from knowledge base entities to their mentions in Wikipedia, the annotation
contains noise.
For example, ``Karl Marx'' is annotated with \texttt{person/author}, \texttt{person/politician} and \texttt{person} and so is every mention of him based on distant supervision 
which is unlikely to be true.
To reduce noise, we
sacrifice some granularity in the \sclass es.
We only use  the 34 \emph{parent} \sclass es in the FIGER hierarchy that
have instances in WIKI-PSE; see \tabref{types}.
For example, we leave out \texttt{person/author} and \texttt{person/politician} and just use \texttt{person}.
By doing so, mentions of nouns are rarely ambiguous with
respect to  \sclass\,
and we still have a reasonable number
of \sclass es (i.e., 34).

The next step is to aggregate all  \sclass es a surface
form is annotated with.  Many surface forms are used for
referring to more than one Wikipedia page and, therefore,
possibly to more than one \sclass.  So, by using these
surface forms of nouns\footnote{Linked multiwords are
treated as single tokens.}, and their
aggregated derived \sclass es, we build our dataset of
\emph{words} and \emph{\sclass es}.  See \figref{apple} for
``apple'' as an example.

We differentiate linked mentions by enclosing them with ``@'', e.g.,  ``apple'' $\rightarrow$ ``@apple@''. 
If 
the mention of a noun is not
linked to a Wikipedia page, then it is not changed, e.g.,
its surface form remains ``apple''.
This prevents conflation of 
\sclass-annotated mentions with unlinked mentions.

For the corpus, we include only sentences with at least one
annotated mention resulting in 550 million tokens --
an appropriate size for embedding learning. 
By lowercasing the corpus and setting the minimum frequency to 20,
the vocabulary size is $\approx$500,000.
There are $\approx$276,000 annotated words in the vocabulary,
each with $>=1$ \sclass es.
In total, there are $\approx$343,000 word/\sclass\ pairs,
i.e., words have 1.24 \sclass es on average. 

For efficiency, 
we select a subset of  words for WIKI-PSE.
We first add all multiclass words (those with more than one
\sclass) to the dataset, divided randomly into train and test (same size). 
Then, we add a random set with the same size from single-class words,
divided randomly into train and test  (same size). 
The resulting train and test sets have the size of 44,250 each, with an equal number of 
single and multiclass words. 
The average number of \sclass es per word is 1.75.

\section{Probing for Semantic Classes in Word Embeddings}
\seclabel{overall}
We investigate
embeddings by probing: Is the information we care about  
available in a word $w$'s embedding?
Specifically, we probe
for
\sclass es:
Can the information whether $w$ belongs to a specific \sclass\
be obtained from its embedding?
The probing method we use
 should be: (i) simple with only the word embedding as input, so that we do not conflate the 
quality of embeddings with other confounding factors like
quality of context representation (as in  WSD);
(ii) supervised with enough training data so that we can learn strong and non-linear classifiers to extract meanings from  embeddings;
(iii)  agnostic to the model architecture that the word embeddings are trained with.

WIKI-PSE, introduced in \secref{dataset}, provides a text
corpus and annotations for setting up probing methods
satisfying (i) -- (iii).
We now describe
the other elements of our experimental setup:
word and sense representations, probing tasks and classification models.

%We can also ask the question: if the embedding of a word $w$
%contains all information in
%principle, can we build machinery that successfully extracts
%it from the embedding? 
%A specialization of this question is:
%if the embedding of a word $w$ faithfully represents all
%meanings of $w$,
%can we build machinery that successfully identifies which
%sense is used in a particular context? This is the task of
%word sense disambiguation.
%
%Clearly, disambiguation is a very important task. But
%separating the two core questions of (i) the quality of the
%representation / embedding and (ii) our ability to build
%machinery that can extract the information seems promising
%to us because it neatly factorizes a complex problem into
%two conceptually clearer problems.

\subsection{Representations of Words and Senses}
We run word embedding models like \textsc{word2vec}
on WIKI-PSE to get embeddings for all  words in the
corpus, including
special common and proper nouns  like ``@apple@''.

We also 
learn an embedding for each \sclass\ of a word, e.g., 
one embedding for  ``@apple@-food'' and one for  ``@apple@-organization''. 
To do this, each  annotated mention of a noun (e.g.,
``@apple@'') is replaced with a word/\sclass\ token
corresponding to its annotation (e.g., with
``@apple@-food'' or ``@apple@-organization'').
These word/\sclass\ embeddings correspond to sense
embeddings in other work.

Finally, we create an alternative word embedding for an ambiguous word
like ``@apple@'' by
aggregrating its word/\sclass\ embeddings by summing them:
$\vec{w} = \sum_i \alpha_i \vec{w_{c_i}}$ where
$\vec{w}$ is the aggregated word embedding and
the $\vec{w_{c_i}}$ are the word/\sclass\ embeddings.
We consider two
aggregations:
\begin{itemize}
\item For \textbf{uniform} sum, written as \textbf{\uniform},
we set $\alpha_i$ = 1. So a word is represented as  the
sum of its sense (or \sclass) embeddings; e.g., the representation
of ``apple'' is the sum of its organization and food
\sclass\ vectors.
\item For \textbf{weighted} sum, written as \textbf{\weighted}, 
we
set
$\alpha_i = \mbox{freq}(w_{c_i}) / \sum_j \mbox{freq}(w_{c_j})$, i.e.,
the relative frequency of word/\sclass\ $w_{c_i}$ in
mentions of the word $w$.
So a word is represented as the \emph{weighted}
sum of its sense (or \sclass) embeddings;
e.g., the representation
of ``apple'' is the weighted sum of its organization and food
\sclass\ vectors where the organization vector receives a
higher weight since it is more frequent in our corpus.
\end{itemize}

\uniform\ is common
in  multi-prototype embeddings, cf.\ \cite{rothe2017autoextend}. 
\weighted\ is also motivated by 
prior work \cite{arora18tacl}.
Aggregation allows us to investigate
the reason for poor performance of single-vector
embeddings. Is it a problem that a
single-vector representation is used as the multi-prototype literature claims? 
Or are single-vectors
in principle sufficient, but the way sense
embeddings are aggregated in a single-vector
representation (through an embedding algorithm, through
\uniform\ or through \weighted)
is critical.

%In contrast to multi-prototype embeddings,  we learn
%embeddings for \sclass es of words instead of induced or
%dictionary senses. 
%Also, the WIKI-PSE  corpus is annotated with \sclass es 
%and 

\begin{figure}
\begin{center}
\includegraphics[scale=0.87]{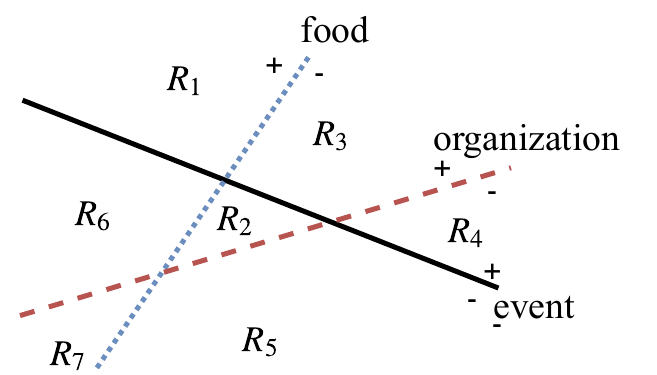}
\caption{A 2D  embedding space with three \sclass es (food, organization and event).
A line divides positive and negative regions of each \sclass.
Each of the seven $R_i$ regions corresponds to a subset of 
\sclass es.}
\figlabel{space}
\end{center}
\end{figure}

\enote{ea}{
  - Figure 1: venn diagrams could be more useful, as in the current drawing
it is not clear which parts are food are which are not food (same for
organization or event).

}
\subsection{Probing Tasks} 
\seclabel{classifiers}
The first task is to probe for \sclass es.
We train, for each \sclass, a binary classifier
that takes an embedding as input   and
predicts membership in the \sclass.
An ambiguous word like ``@apple@'' belongs to
multiple \sclass es, so each of several
different binary classifiers should diagnose it as being
in its \sclass. 
%, e.g., ``food'' and ``organization''
%for ``apple''.
How well this type of probing for \sclass es works in practice
is one of our key questions: can \sclass es be
correctly encoded in embedding space?

\figref{space} shows a 2D embedding space:
each point is assigned to a subset of the
three \sclass es, e.g., ``@apple@'' is in the region
``+food $\cap$ +organization $\cap$ -event'' and ``@google@'' in the region ``-food $\cap$ +organization $\cap$ -event''.

The second probing task
predicts whether an embedding represents an
unambiguous (i.e.,
one \sclass) or an ambiguous (i.e., multiple \sclass es) word.
Here, we do not look for any specific meaning in an
embedding, but  assess whether it is an encoding of multiple different meanings or not.
High accuracy of this
classifier
would
imply that ambiguous and unambiguous words are
distinguishable in the embedding space.

\subsection{Classification Models}
Ideally, we would like to have linearly separable spaces 
with respect to \sclass es -- presumably embeddings from
which information can be effectively extracted by such a
simple mechanism are better.
However, this might not be the case considering the
complexity of the space:
non-linear models may  detect \sclass es
more accurately. 
Nearest neighbors computed by cosine similarity
are frequently used to classify and analyze embeddings, so
we consider them as well.
Accordingly, we experiment with three classifiers:
 (i) logistic regression (LR); 
 (ii) multi-layer perceptron (MLP) with one 
 hidden and a final ReLU layer; and 
 (iii) KNN: K-nearest neighbors.

\section{Experiments}
\seclabel{setup} \textbf{Learning embeddings}.  Our method
is agnostic to the word embedding model.  Therefore, we
experiment with two popular similar embedding models: (i)
SkipGram (henceforth \textbf{\textsc{skip}})
\cite{mikolov2013efficient}, and (ii) Structured SkipGram
(henceforth \textbf{\textsc{sskip}})
\cite{ling15embeddings}.  \textsc{sskip} models word order while \textsc{skip} is a
bag-of-words model.  We use \textsc{wang2vec}
\cite{ling15embeddings} with negative sampling for training
both models on WIKI-PSE.
For each model, we try four embedding sizes: \{100, 200,
300, 400\} using identical hyperparameters: negatives$=$10,
iterations$=$5, window$=$5.

\begin{table}[h]
\footnotesize
\centering{
\begin{tabular}
{c|c|r|lcc}             
emb & size &ln & LR & KNN & MLP \\
\hline 

\hline 
\multirow{4}{*}{\scriptsize{\begin{tabular}{c}
\textsc{skip} \\ word \end{tabular}}}
& 100 & 1 &  .723 & .738 & .773 \\
& 200 & 2 & .740 & .734 & .786 \\
& 300 & 3 & .745 & .730 & .787 \\
& 400 & 4 & .747 & .727 & .786 \\
\hline 
\multirow{4}{*} {\scriptsize{\begin{tabular}{c}
\textsc{skip} \\ \weighted \end{tabular}}}
& 100 & 5 & .681 & .727 & .752 \\
& 200 & 6 & .695 & .721 & .756 \\
& 300 & 7 & .699 & .728 & .752 \\
& 400 & 8 & .702 & .711 & .753 \\
\hline 
\multirow{4}{*} {\scriptsize{\begin{tabular}{c}
\textsc{skip} \\ \uniform  \end{tabular}}}
& 100 & 9 & .787 & \textbf{.783} & .830 \\
& 200 & 10 & .797 & .773 & .833 \\
& 300 & 11 & .800 & .765 & .832 \\
& 400 & 12 & \textbf{.801} & .758 & \textbf{.834} \\

\hline \hline 
\multirow{4}{*}{\scriptsize{\begin{tabular}{c}
\textsc{sskip} \\ word \end{tabular}}}
 & 100 & 13 & .737 & .749 & .785 \\
 & 200 & 14 & .754 & .745 & .793 \\
 & 300 & 15 & .760 & .741 & .797 \\
 & 400 & 16 & .762 & .737 & .790 \\
\hline 
\multirow{4}{*} {\scriptsize{\begin{tabular}{c}
\textsc{sskip} \\ \weighted \end{tabular}}}
& 100 & 17 & .699 & .733 & .762 \\
 & 200 & 18 & .710 & .726 & .764 \\
 & 300 & 19 & .714 & .718 & .767 \\
 & 400 & 20 & .717 & .712 & .763 \\
\hline 
\multirow{4}{*} {\scriptsize{\begin{tabular}{c}
\textsc{sskip} \\ \uniform \end{tabular}}}
 & 100 & 21 & .801 &  \textbf{.783}& .834 \\
 & 200 & 22 & .809 & .767 & .840 \\
 & 300 & 23 & .812 & .755 & .842 \\
 & 400 & 24 &\textbf{.814} & .747 &\textbf{ .844} \\
\hline \hline 
random & -- & -- & .273& -- & --\\

\end{tabular}
}
\caption{$F_1$ for \sclass\ prediction. emb: embedding,
 \uniform\ (resp.\ \weighted): uniform (resp.\ weighted) sum of word/\sclass es.
ln: line number. 
Bold: best $F_1$ result per column and embedding model (\textsc{skip} and \textsc{sskip}).
}
\tablabel{results}
\end{table}

\begin{figure*}[h]
\begin{subfigure}{.5\textwidth}
  \centering
  \includegraphics[width=1\textwidth]
  {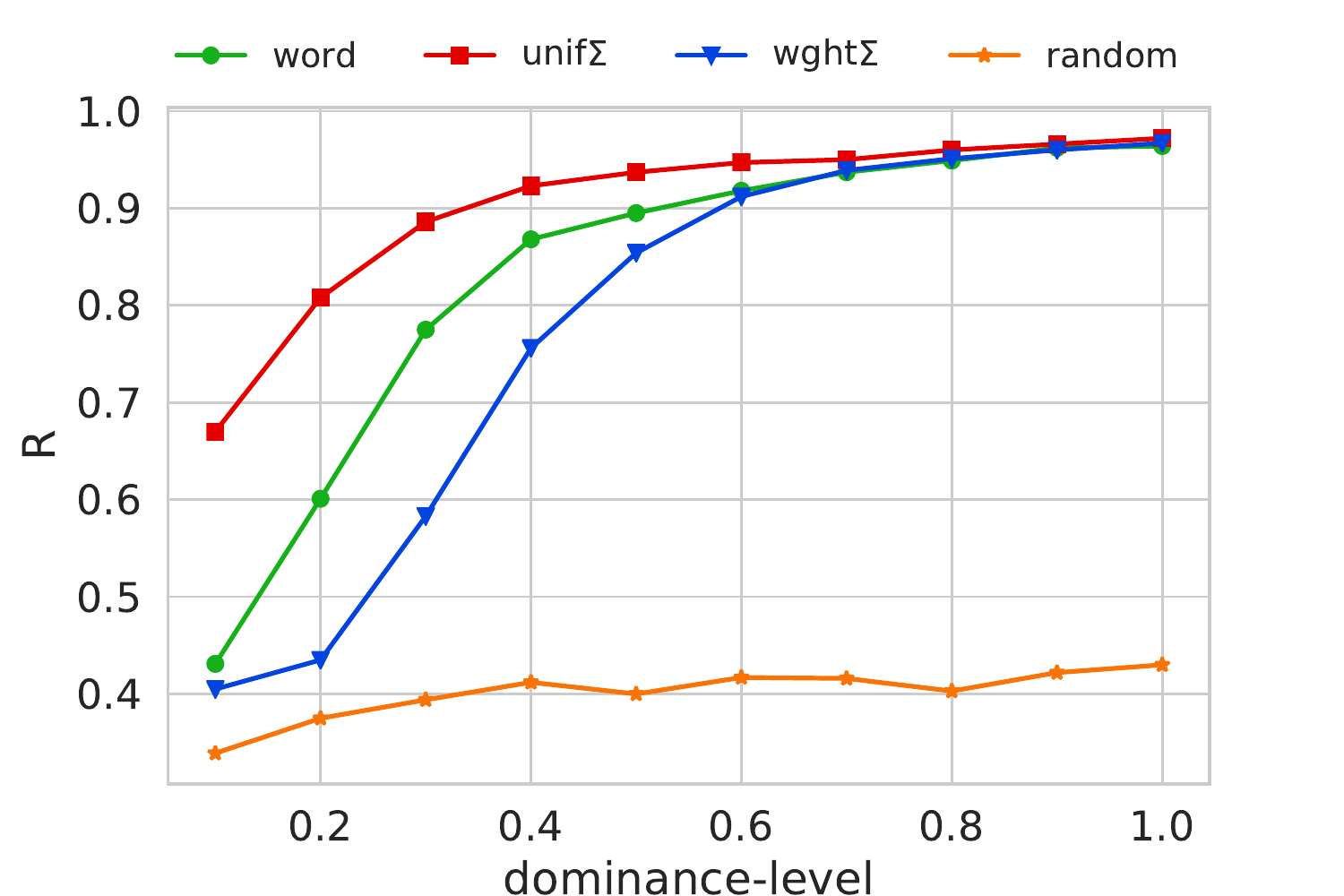}
  \caption{}
  \figlabel{dominance}
\end{subfigure}
~
%\begin{subfigure}{.5\textwidth}
%  \centering
%  \includegraphics[width=180pt,height=110pt]
%  {plots/frequency-level}
%  \caption{}
%  \figlabel{freq}
%\end{subfigure}
%~
\begin{subfigure}{.5\textwidth}
  \centering
  \includegraphics[width=1\textwidth]
  {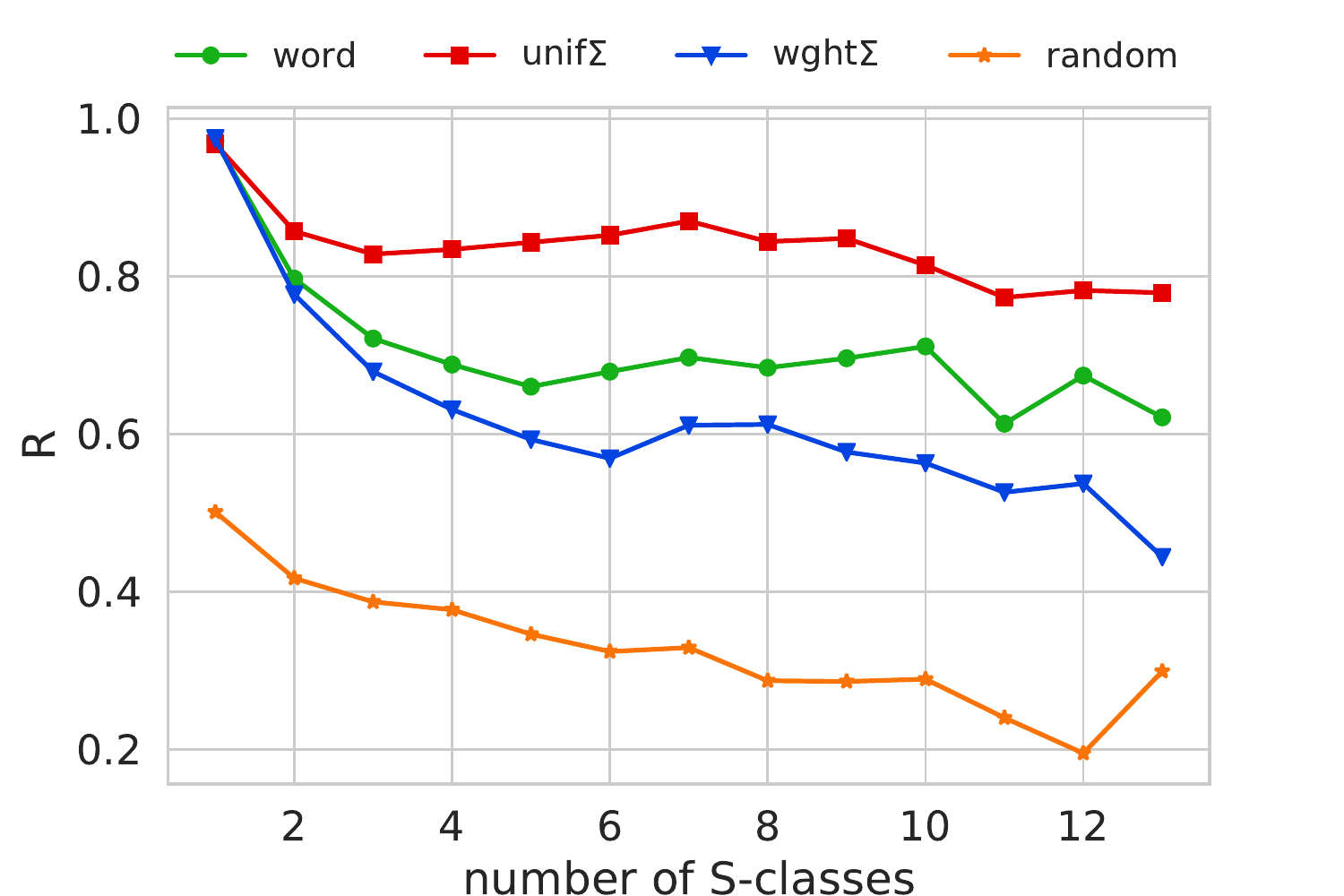}
  \caption{}
  \figlabel{classes}
\end{subfigure}
%~
%\begin{subfigure}{.5\textwidth}
%  \centering
%  \includegraphics[width=180pt,height=110pt]
%  {plots/typicality-level}
%  \caption{}
%  \figlabel{typicality}
%\end{subfigure}
\caption{Results of \sclass\ prediction as a function of two
  important factors:
  dominance-level and number of \sclass es}
\figlabel{mlpfigures}
\end{figure*}

\subsection{\sclass\ Prediction}
\tabref{results} shows 
results on \sclass\ prediction for word, \uniform\ and \weighted\
embeddings trained using \textsc{skip} and \textsc{sskip}. 
Random is a simple baseline that
randomly assigns to a test example
each \sclass\
according
to its prior probability (i.e., proportion in train).

We train classifiers with Scikit-learn \cite{scikit-learn}.
Each classifier is an independent binary predictor for one  \sclass.
We use the global metric of micro $F_1$ over all test
examples and over all \sclass\ predictions.
We see the following trends in our results.

MLP  is consistently better than LR or KNN. Comparing MLP and LR reveals that the space is not linearly separable with respect to the \sclass es. 
This means that linear classifiers are insufficient for
semantic probing: \emph{we should use models for probing that are
more powerful than linear}.

Higher dimensional embeddings perform better  for MLP
and LR, but worse for KNN. 
We do further analysis by
counting the number $k$ of unique \sclass es in the top 5 nearest
neighbors for word embeddings; $k$ is 1.42 times larger
for embeddings of dimensionality 400
than 200.
Thus, \emph{more dimensions results in more diverse neighborhoods
and more randomness}.
We explain this by the increased degrees of freedom in a
higher dimensional space: idiosyncratic properties of words
can also be represented given higher capacity and so
similarity in the space is more influenced by
idiosyncracies, not by general properties like semantic classes.
Similarity datasets tend to only test the majority
sense of words \cite{gladkova16}, and that is perhaps why
similarity results usually do not follow the same trend
(i.e., higher dimensions improve results).
See \tabref{similarity} in Appendix for results on selected similarity
datasets.

\textsc{sskip} performs better than \textsc{skip}.
The difference between the two is that \textsc{sskip} models
word order.
Thus, we conclude that \emph{modeling word order is important for a robust representation}.
This is in line with the more recent \textsc{fasttext} model with word order that
outperforms prior work \cite{mikolov2017advances}.

We now compare word embeddings,
\uniform, and \weighted. Recall that the sense
vectors of a word have equal weight in \uniform\ and are
weighted according to their frequency in \weighted. The
results for word embeddings (e.g., line 1) are between those of
\uniform\ (e.g., line 9) and \weighted\ (e.g., line 5). This
indicates that
their weighting of sense vectors is somewhere between the
two extremes of \uniform\ and \weighted. Of course, word
embeddings are not computed as an explicit weighted sum of
sense vectors, but there is evidence that they are implicit
frequency-based weighted sums of  meanings or
concepts \cite{arora18tacl}.

The ranking \uniform\ $>$ word embeddings $>$
\weighted\ indicates how well individual sense vectors are represented
in the aggregate word vectors
and how well they can be ``extracted'' by a classifier in
these three representations.
Our prediction task
is designed to find \emph{all} meanings of a word, including
rare senses.
\uniform\ is designed to give relatively high weight to rare
senses, so it does well on the prediction task.
\weighted\
 and word embeddings give low weights to rare senses and
 very high weights to frequent senses, so
 the rare senses can be ``swamped'' and difficult to extract by classifiers
 from the embeddings.

\begin{table}[h]
\footnotesize
\centering
\begin{tabular}
{l||lcc}             
emb & LR & KNN & MLP \\
\hline\hline
word & .711 & .605 & .715 \\
\weighted & .652 & .640 & .667 \\
\uniform  & \textbf{.766} & \textbf{.709} & \textbf{.767} \\
\hline
\textsc{glove}(6B)     & .667 & .638 & .685 \\
\textsc{fasttext}(Wiki)        & .699 & .599 & .697 
%\textsc{elmo}(Ch-CNN)     & .510 & .479 & .481 \\
\end{tabular}
\caption{$F_1$ for \sclass\  prediction on the subset of
  WIKI-PSE
whose vocabulary is shared with \textsc{glove} and
\textsc{fasttext}.
Apart from using a subset of WIKI-PSE, this is the same
setup as in \tabref{results}, but here we compare 
word, \weighted, and \uniform\ with public \textsc{glove} and \textsc{fasttext}.
}
\tablabel{results-onewords}
\end{table}

\textbf{Public embeddings.}
To give a sense on how well public embeddings, trained on much larger data, do on \sclass\ prediction in WIKI-PSE,
we use 300d \textsc{glove} embeddings trained on 6B tokens\footnote{https://nlp.stanford.edu/projects/glove/} from Wikipedia and Gigaword and \textsc{fasttext} Wikipedia word embeddings.\footnote{https://fasttext.cc/docs/en/pretrained-vectors.html}
We create a subset of the WIKI-PSE dataset by keeping only single-token
words that exist in the two embedding vocabularies.
The size of the resulting dataset is 13,000 for  train and test each;
the average number of \sclass es per word is 2.67.

\tabref{results-onewords} shows  results and compares with our different \textsc{sskip} 300d embeddings.
There is a clear performance gap between the two off-the-shelf embedding models and \uniform,
indicating that training on larger text does not necessarily help for prediction of rare meanings.
This table also confirms \tabref{results} results with
respect to comparison of learning model (MLP, LR, KNN) and
embedding model (word, \weighted, \uniform).
Overall, the performance drops compared to the results in \tabref{results}.
Compared to the WIKI-PSE dataset, this subset has fewer (13,000 vs.\ 44,250)
training examples, and a larger number of labels per example (2.67 vs.\ 1.75).
Therefore, it is a harder task.

\subsubsection{Analysis of Important Factors}
\seclabel{factors}
We analyze the performance with respect to multiple factors
that can influence the quality of
the representation
of \sclass\ $s$
in the 
embedding of word $w$: dominance, number of
\sclass es, frequency and typicality.
We discuss the first two here and the latter two in the Appendix \secref{factorsAppendix}.
These factors are similar to those affecting WSD systems \cite{pilehvar2014psuedoword}.
We perform this analysis for MLP classifier on \textsc{sskip} 400d embeddings.
We compute the recall for various conditions.\footnote{Precision for these cases is not defined. This is similarly applied in WSD \cite{pilehvar2014psuedoword}.}

\textbf{Dominance} of the \sclass\ $s$ for word $w$ is defined as
the percentage of the occurrences of $w$ where its labeled \sclass\ is $s$.
\figref{dominance} shows for each dominance level what
percentage of \sclass es of that level were correctly
recognized by their binary classifier. 
For example,
0.9 or 90\% of
\sclass es of words with
dominance level 0.3 were
correctly recognized by the corresponding \sclass's binary
classifier for
\uniform\ ((a), red curve).
Not surprisingly,
more dominant meanings are represented and
recognized better.

We also see that word embeddings represent non-dominant
meanings better than \weighted, but worse than \uniform.
For word embeddings, the
performance drops sharply for dominance $<$0.3.
For \weighted, the sharp drops happens earlier, at dominance $<$0.4.
Even for \uniform, there is a (less sharp) drop -- 
this is due to other factors like frequency and not due to
poor representation of less dominant \sclass es (which all
receive equal weight for \uniform).

The \textbf{number of \sclass es} of a word can influence  
the quality of meaning extraction from its embedding.
\figref{classes} confirms our expectation:
It is  easier to extract a meaning from a word embedding that encodes fewer meanings. 
For words with only one \sclass, the result is best.
For ambiguous words, performance drops but this is less of an
issue for \uniform. For word embeddings (word), performance remains in the range  0.6-0.7 for more than 3
\sclass es which is lower than \uniform\ but higher than \weighted\ by around 0.1.

%\begin{figure}[h!]
%\centering
%\includegraphics[scale=0.35]{plots/skewed_numtypes.png}
%\caption{The average number of unique \sclass es in the nearest neighbors of words with two classes,
%in different dominance level.}
%\figlabel{uniqTypesNeighbors}
%\end{figure}
%
%\subsubsection{What does happen when classes of a word become balanced?}
%Here, we analyze the space of word embeddings with multiple \sclass es as the class distribution gets more balanced. 
%In \figref{uniqTypesNeighbors}, we show that for two-\sclass\  words,
%the average number of unique classes in the top five nearest neighbors increases as the dominance level increases.
%The dominance-level of 0.4 is basically where the two classes are almost equally frequent.
%As the two classes move towards equal importance,
%their word embeddings move towards a space with more diversity.

% We discuss dependencies between the four factors
% (dominance, number of
% \sclass es, frequency and typicality)
% in Supplementary.

\subsection{Ambiguity Prediction}
\seclabel{ambpred}
We now investigate if
a classifier can predict 
whether a word is ambiguous or not, based on the word's embedding.
We divide the WIKI-PSE dataset into two groups:
unambiguous (i.e., one \sclass)
and ambiguous (i.e., multiple \sclass es). 
LR, KNN and MLP are trained on the training set and applied to the words in test.
The only input to a classifier is the embedding; the
output is binary: one
\sclass\ or multiple \sclass es.
We use 
\textsc{sskip} word embeddings
(dimensionality 400)
and L2-normalize all vectors
before classification.
% Baselines include a majority baseline (MAJORITY) and a LR baseline that uses the
% word frequency
% as single feature (FREQUENCY).
As a baseline, we use the
word frequency
as single feature (FREQUENCY) for LR classifier.

\begin{table}[h]
\footnotesize
\centering
\begin{tabular}{l|l|ccc}
model & LR &  KNN & MLP\\
\hline
% MAJORITY & 56.5 & -  & -\\
FREQUENCY & 64.8 & - & -\\
\hline 
word &  77.9  & 72.1 & 81.2 \\
%SKIP-word & no & 76.3 & 72.1 & 80.6 \\
\hline
\weighted & 76.9 & 69.2 & 81.1\\
%\textbf{w/s}(weighted) & no &  70.6 & 69.2 & 80.9 \\
\hline
\uniform & 96.2 & 72.2 & 97.1 \\
%\textbf{w/s}(uniform) & no & 98.7 & 72.2 & 98.7 \\
\end{tabular}
\caption{Accuracy for predicting ambiguity}
\tablabel{ambiguitytab}
\end{table}

\begin{figure}[h]
  \centering
   \includegraphics[width=240pt]{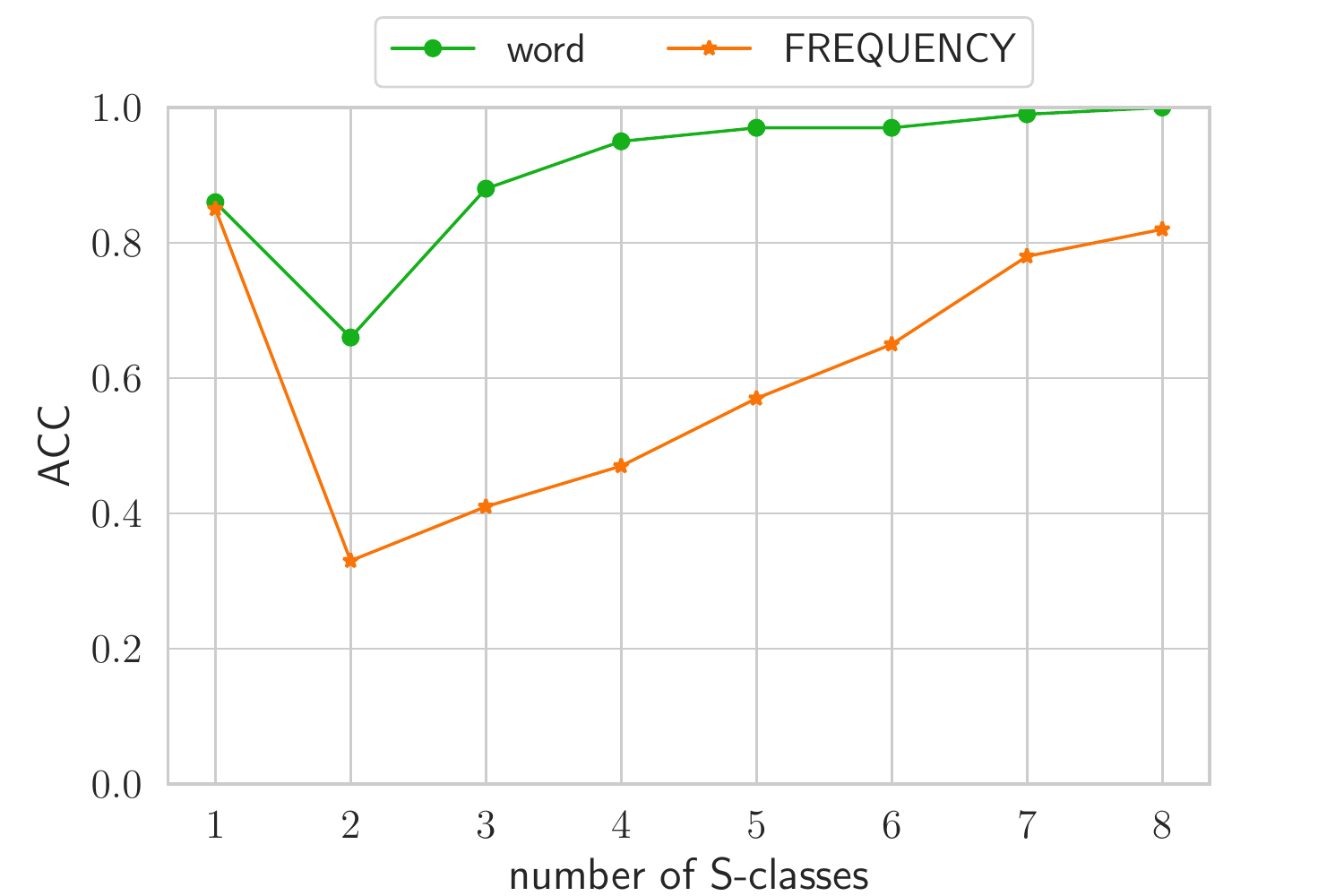}
\caption{Accuracy of word embedding and FREQUENCY for predicting
  ambiguity as a function of number of S-classes, using MLP classifier.}
\figlabel{ambiguity}
\end{figure}

%Vector normalization improves the performance, indicating that the magnitude of the vectors is not relevant here and only the direction matters. 

\tabref{ambiguitytab} shows overall accuracy and
\figref{ambiguity} accuracy as a function of number of
\sclass es. Accuracy of standard word embeddings is clearly above the 
baselines, e.g., 81.2\% for MLP and 77.9\% for LR
compared to 64.8\% for FREQUENCY.
The figure shows that the decision
becomes easier with increased ambiguity (e.g.,
$\approx$100\% for 6 or more \sclass es). It makes sense
that a highly ambiguous word is more easily identifiable
than a two-way ambiguous word.
MLP accuracy for \uniform\ is close
to 100\%. We can again attribute this to the fact that
rare senses are better represented in
\uniform\ than in
regular word embeddings, so the ambiguity classification is easier.

KNN results are worse than LR and MLP.
This indicates that
similarity is not a good indicator of degree of ambiguity:
words with similar degrees of ambiguity do not seem to be neighbors of
each other. This observation also points to 
an explanation for why the classifiers achieve such high
accuracy. We saw before that \sclass es can be identified
with high accuracy. Imagine a multilayer architecture that
performs binary classification for each \sclass\ in the
first layer and, based on that, makes the ambiguity decision
based on the number of \sclass es found. LR and MLP seem to
approximate this architecture. Note that this can only work
if the individual \sclass es are recognizable, which is not
the case for rare senses in regular word embeddings.

In Appendix \secref{ambPredAppendix}, we
show top predictions for ambiguous and unambiguous words.

\subsection{NLP Application Experiments}
Our primary goal is to probe meanings in word embeddings without confounding 
factors like contextual usage. 
However, to give insights on how our probing results relate to NLP tasks, we evaluate 
our embeddings when used to represent word
tokens.\footnote{For the embeddings used in this experiment, if there are versions
  with and without ``@''s, then we average the two;
e.g.,  ``apple'' is the average of ``apple'' and ``@apple@''.} 
Note that our objective here is not to improve over other baselines, but to perform analysis.

We select mention, sentence and sentence-pair classification datasets.
For mention classification, we adapt \newcite{shimaoka17eacl}'s setup:\footnote{https://github.com/shimaokasonse/NFGEC} training, evaluation (FIGER dataset) and implementation.
The task is to predict the contextual fine-grained types of entity mentions.
We lowercase the dataset to match the vocabularies of \textsc{glove}(6B), \textsc{fasttext}(Wiki)
and our embeddings.
For sentence and sentence-pair classifications, we use the SentEval\footnote{https://github.com/facebookresearch/SentEval} \cite{conneau2018senteval} setup for four datasets:
MR \cite{Pang05MR} (positive/negative sentiment prediction for movie reviews) , 
CR \cite{hu04CR} (positive/negative sentiment prediction for product reviews), 
SUBJ \cite{Pang04subj} (subjectivity/objectivity prediction) and  
MRPC \cite{dolan04mrpc} (paraphrase detection).
We average embeddings to encode a sentence.

\begin{table}[h]
\addtolength{\tabcolsep}{-1pt}
\footnotesize
\centering
\begin{tabular}
{l|ccccc}          
emb & MC & CR & MR & SUBJ  & MRPC\\
\hline 
word & 64.6 & 70.4 & 71.4 & 89.2 & 71.3 \\
\weighted & 65.4 & 72.3 & 72.0 & 89.4  & 71.5 \\
\uniform & 61.6 & 69.1 & 68.8 & 87.9 & 71.3 \\
%\textsc{sskip}(w/s-avg) & 71.7 & 71.6 & 89.3& 71.9\\

\hline
\textsc{glove}(6B)       & 58.1& 75.7  & 75.2 & 91.3  & 72.5  \\
\textsc{fasttext}(Wiki) & 55.5&  76.7 & 75.2 & 91.2  & 71.6 \\
%\textsc{glove}(840B)   & 65.6&  76.9 & 78.36 & 91.2  & 73.1  \\
%\textsc{elmo}(Ch-CNN)            & 50.1 &  74.5 & 78.0 & 90.2  & 72.1  \\
%Glove(6B)-shared    & 76.6 & 74.3 & 90.8  &  69.7 \\
%FSTTXT-shared  & 75.8 & 74.8 & 91.3  & 71.1 \\

\end{tabular}
\caption{Performance of the embedding models on five NLP tasks}
\tablabel{tasks}
\end{table}

\tabref{tasks} shows  results.
For MC,
performance of embeddings is ordered:
\weighted\ $>$ word $>$ 
\uniform.  This is the opposite of
the ordering in \tabref{results} where \uniform\ was the
best and \weighted\ the worst.  
The models with more weight on frequent meanings perform
better  in this task, likely because the dominant \sclass\ is mostly what is needed. 
In an error analysis, we found many cases where
mentions have one major sense and some minor senses;
e.g., \uniform\ predicts ``Friday''  to be ``location'' in
the context ``the U.S. Attorney's Office announced
Friday''. Apart from the major \sclass\ ``time'',
``Friday'' is also a mountain
(``Friday Mountain'').
\uniform\ puts the same weight on ``location'' and ``time''.
\weighted\ puts almost no weight on ``location'' and
correctly predicts ``time''.
Results for the four other datasets are  consistent: the
ordering is the same as for MC.

%This again indicates that in this kind of target task, embeddings with better representation of rare meanings are worse in terms of overall performance. 

\section{Discussion and Conclusion}
We quantified how well multiple meanings are represented in
word embeddings.  We did so by designing two probing tasks,
\sclass\ prediction and ambiguity prediction.  We applied
these probing tasks on WIKI-PSE, a large new resource for
analysis of ambiguity and word embeddings.  We used \sclass
es of Wikipedia anchors to build our dataset of
word/\sclass\ pairs. We view \sclass es
as corresponding to senses.

A summary of our findings is as follows.
(i) We can build a classifier that,
with high accuracy, correctly predicts whether an
embedding represents an ambiguous or an unambiguous word.
(ii) We show
 that semantic classes are
recognizable in embedding space -- a novel result as far as
we know for a real-world dataset 
-- and much better with a nonlinear classifier than a linear one. 
(iii) The standard word embedding models learn embeddings
 that capture multiple meanings in a single vector well --
 if the meanings are frequent enough. 
(iv) Difficult cases of ambiguity --  rare word senses or 
 words with numerous senses -- are better captured when the
  dimensionality of the embedding space is increased. But
  this comes at a cost -- specifically, cosine similarity
 of embeddings (as, e.g., used by KNN, \secref{ambpred}) becomes less predictive of \sclass.
(v)
 Our diagnostic tests show that a uniform-weighted sum of the senses of a
 word $w$ (i.e., \uniform) is a high-quality representation of all senses of $w$
 -- even if the word embedding of $w$ is not. This suggests
again  that the main problem is not ambiguity per se, but
rare senses.
(vi) Rare senses are badly represented if we use explicit
frequency-based weighting of meanings
(i.e., \weighted)
compared to word embedding learning models like SkipGram.

To relate these findings to sentence-based applications, we experimented 
with a number of public classification datasets. 
Results suggest that embeddings with frequency-based
weighting of meanings work better for these tasks.
Weighting all meanings equally means that a highly dominant
sense (like ``time'' for ``Friday'') is severely
downweighted. This indicates that currently used tasks
rarely need rare senses -- they do fine if they have only
access to frequent senses.
However, to achieve
high-performance natural language understanding at the human
level, our models also need to be able to have access to
rare senses -- just like humans do.  
We conclude that we need harder NLP tasks for
which performance depends on rare as well as frequent
senses. Only then will we be able to show the benefit of
word representations that represent rare senses accurately.

%These findings give us a better understanding of word
%embeddings and how we embeddings deal with word ambiguity.

%We are planning to exploit these insights
%for tackling ambiguity in NLP applications. 
%One simple
%strategy could be to first
%identify ambiguous words (based on their embeddings) and
%then difficult ambiguous words.
%If we can devise a method for doing this reliably,
%then we can
%compute regular embeddings for unambiguous and
%unproblematic ambiguous words and reserve complex
%methods (such as context clustering and learning uniform sum of cluster embeddings) for hard ambiguous
%words.
%
%As another line of future work, we can extend WIKI-PSE to other languages and test our findings on new embedding spaces. 

\section*{Acknowledgments}
We are grateful for the support of the European Research
Council (ERC \#740516) and UPV/EHU 
(excellence research group) for this work.
Next, we thank all the anonymous reviewers 
their detailed assessment and helpful comments.
We also  appreciate the insightful discussion with Geoffrey J. Gordon, Tong Wang, and other members
of Microsoft Research Montr\'eal.

\bibliography{main}
\bibliographystyle{acl_natbib}

\appendix
\section{Analysis of important factor: more analysis}
\seclabel{factorsAppendix}

\begin{figure*}[h!]
\begin{subfigure}{.5\textwidth}
  \centering
  \includegraphics[width=215pt,height=140pt]
  {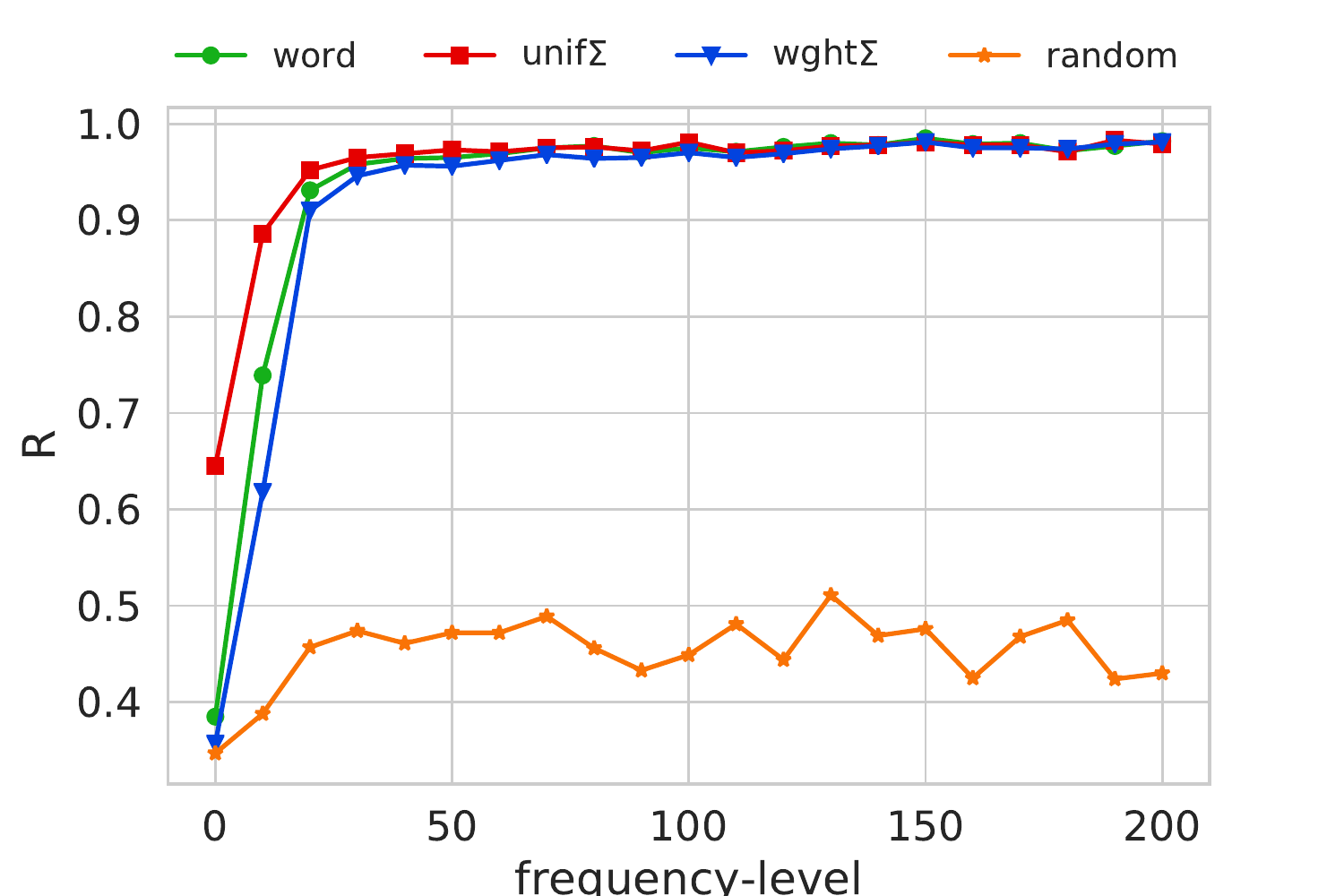}
  \caption{}
  \figlabel{freq}
\end{subfigure}
~
\begin{subfigure}{.5\textwidth}
  \centering
  \includegraphics[width=215pt,height=140pt]
  {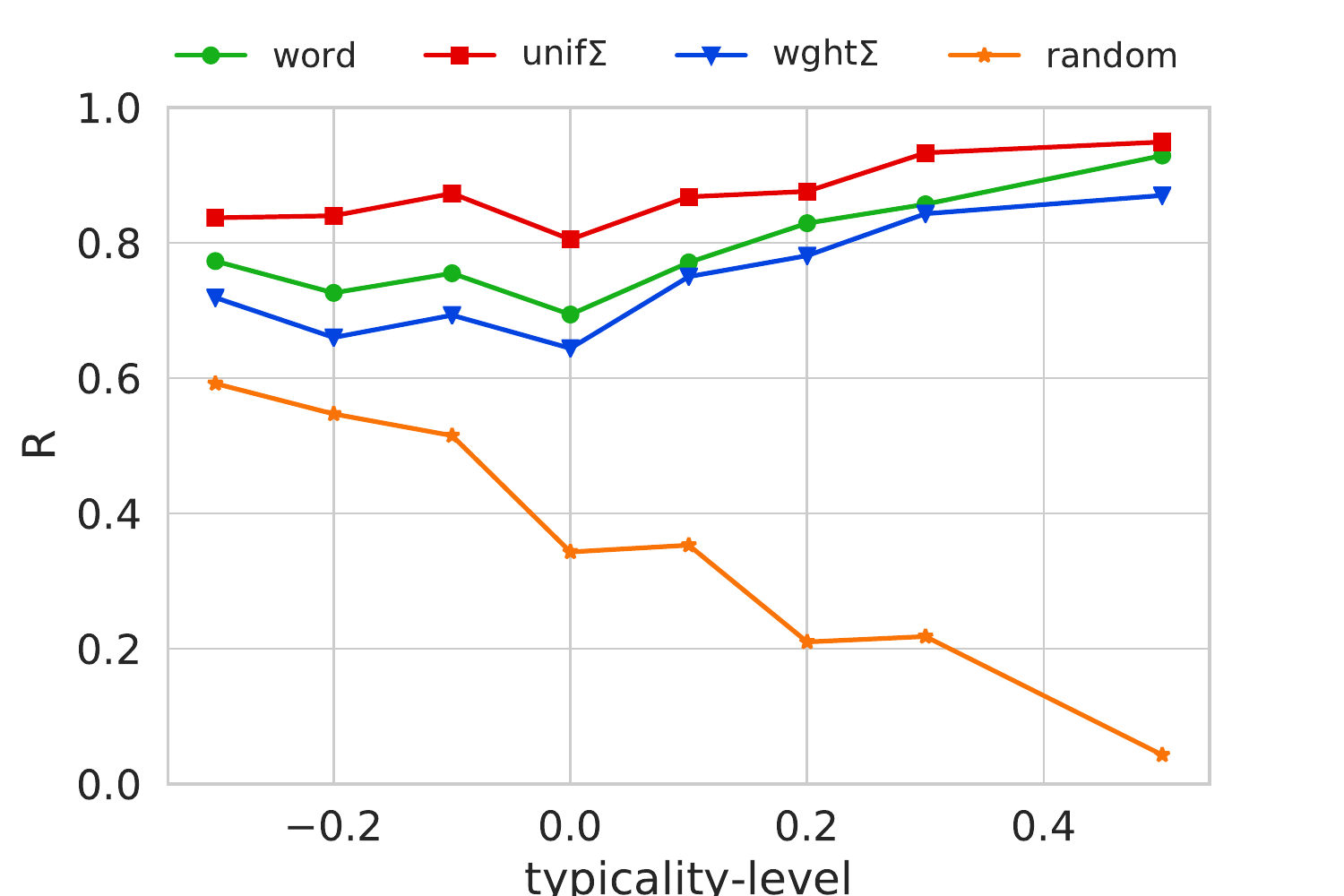}
  \caption{}
  \figlabel{typicality}
\end{subfigure}
~
\caption{Results of word, uniform and weighted word/\sclass\  embeddings for two other important factors: 
frequency and typicality of \sclass.}
\figlabel{extraFactors}
\end{figure*}

\textbf{Frequency} is defined as the absolute frequency of $s$ in occurrences of $w$. Frequency is important to get good representations and the assumption is that more frequency means better results. 
In \figref{freq}, prediction performance
is shown for a varying frequency-level.
Due to rounding, each level in $x$ includes frequencies $[x-5, x+5]$. 
As expected higher frequency means better results. All embeddings  have high performance when frequency is more than 20, emphasizing that embeddings can indeed represent a meaning well if it is not too rare. 
For low frequency word/\sclass\ es, the uniform sum performs
clearly better than the other models. 
This shows that word and weighted word/\sclass\  embeddings are not good encodings for rare meanings.

\textbf{Typicality} of a meaning for a word is important.
We define the typicality of \sclass\ $s$ for word $w$ as its
average compatibility level with other classes of $w$. We
use Pearson correlation between \sclass es in the training words and 
assign the compatibility level of \sclass es based on that.
In \figref{typicality}, we see that more positive typicality
leads to better results in general.  
Each level in $x$ axis represents $[x-0.05, x+0.05]$.
The \sclass es that have negative typicality are often the frequent ones like ``person'' and ``location''
and that is why the performance is relatively good for them.
%This factor is not as impactful as the other ones.

\begin{table*}
\footnotesize
\centering
\begin{tabular}{l|c|cccccccccc}
model        & size & MEN   & MTurk & RW    & SimLex999 & WS353 & Google & MSR   \\
\hline \hline
\textsc{skip} & 100 & 0.633 & 0.589 & 0.283 & 0.276     & 0.585 & 0.386  & 0.317 \\
\textsc{skip} & 200 & 0.675 & 0.613 & 0.286 & 0.306     & 0.595 & 0.473  & 0.382 \\
\textsc{skip} & 300 & 0.695 & 0.624 & 0.279 & 0.325     & 0.626 & 0.495  & 0.405 \\
\textsc{skip} & 400 & 0.708 & 0.630 & 0.268 & 0.334     & 0.633 & 0.506  & 0.416 \\
\hline
\textsc{sskip} & 100 & 0.598 & 0.555 & 0.313 & 0.272     & 0.559 & 0.375  & 0.349 \\
\textsc{sskip} & 200 & 0.629 & 0.574 & 0.310 & 0.306     & 0.592 & 0.464  & 0.413 \\
\textsc{sskip} & 300 & 0.645 & 0.588 & 0.300 & 0.324     & 0.606 & 0.486  & 0.430 \\
\textsc{sskip} & 400 & 0.655 & 0.576 & 0.291 & 0.340     & 0.616 & 0.491  & 0.431
\end{tabular}
\caption{Similarity and analogy results of our word embeddings on a set of datasets \cite{jastrzebski17}.
The table shows the Spearmans correlation  between the model’s similarities and human judgments.
Size is the dimensionality of the embeddings.
Except for RW dataset, results improve by increasing embeddings size.}
\tablabel{similarity}
\end{table*}

\begin{table*}
\centering{
\footnotesize
\begin{tabular}{l|c|p{2.5in}|c}
word & frequency & senses & likelihood\\
\hline \hline 
@liberty@ & 554 & event, organization, location, product, art, person & 1.0\\ 
@aurora@ & 879 & organization, location, product, god, art, person, broadcast\_program & 1.0\\ 
@arcadia@ & 331 & event, organization, location, product, art, person, living\_thing & 1.0\\ 
@brown@ & 590 & food, event, title, organization, visual\_art-color, person, art, location, people-ethnicity, living\_thing & 1.0\\ 
@marshall@ & 1070 & art, location, title, organization, person & 1.0\\ 
@green@ & 783 & food, art, organization, visual\_art-color,  location, internet-website, metropolitan\_transit-transit\_line, religion-religion, person, living\_thing & 1.0\\ 
@howard@ & 351 & person, title, organization, location & 1.0\\ 
@lucas@ & 216 & art, person, organization, location & 1.0\\ 
@smith@ & 355 & title, organization, person, product, art, location, broadcast\_program & 1.0\\ 
@taylor@ & 367 & art, location, product, organization, person & 1.0 \\
... & & & \\
... & & & \\
@tomáš\_cibulec@ & 47 & person & 0.0\\ 
@judd\_winick@ & 113 & person & 0.0\\ 
@roger\_reijners@ & 26 & person & 0.0\\ 
@patrick\_rafter@ & 175 & person & 0.0\\ 
@nasser\_hussain@ & 82 & person & 0.0\\ 
@sam\_wyche@ & 76 & person, event & 0.0\\ 
@lovie\_smith@ & 116 & person & 0.0\\ 
@calliostomatidae@ & 431 & living\_thing & 0.0\\ 
@joe\_girardi@ & 147 & person & 0.0\\ 
@old\_world@ & 91 & location, living\_thing & 0.0\\
\end{tabular}
}
\caption{The top ten ambiguous words followed by the top unambiguous words based on our  model prediction in Section 5.3.
Each line is a word followed by its frequency in the corpus, its dataset senses and finally our ambiguity prediction likelihood to be ambiguous. 
\tablabel{ambiguityExamples}
}
\end{table*}

\section{What does happen when classes of a word become balanced?}
\seclabel{balancedAppendix}
\begin{figure}
\centering
\includegraphics[scale=0.5]{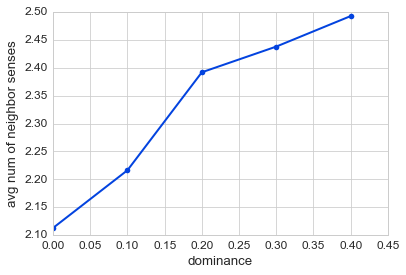}
\caption{The average number of unique semantic classes in the nearest neighbors of words with two classes,
in different dominance level.}
\figlabel{uniqTypesNeighbors}
\end{figure}

Here, we analyze the space of word embeddings with multiple semantic classes as the class distribution gets more balanced. 
In \figref{uniqTypesNeighbors}, we show that for two-class words,
the average number of unique classes in the top five nearest neighbors increases as the dominance level increases.
The dominance-level of 0.4 is basically where the two classes are almost equally frequent.
As the two classes move towards equal importance,
their word embeddings move towards a space with more diversity.

\section{Ambiguity prediction examples}
\seclabel{ambPredAppendix}
In \tabref{ambiguityExamples}, we show some example predicted
ambiguous and unambiguous words based on the word embeddings.

\section{Supersense experiment}
To confirm our results in another dataset, we try supersense annotated Wikipedia of UKP \cite{flekova2016supersense}.
We use their published 200-dimensional word embeddings. 
A similar process as our WIKI-PSE is applied on the annotated corpus to build word/\sclass\  dataset. 
Here, the \sclass es are the supersenses. We consider NOUN categories of words and build datasets for our analysis by aggregating the 
supersenses a word annotated with in the corpus.
Number of supersenses is 26 and train and test size: 27874. 
In \tabref{supersense}, we show the results of ambiguity prediction. 
As we see, we can predict ambiguity using word embeddings with accuracy of 73\%.

\begin{table}
\footnotesize
\centering
\begin{tabular}{l|l|ccc}
model & norm? & LR &  KNN & MLP\\
\hline \hline
MAJORITY & - & 50.0 & -  & -\\
FREQUENCY& - & 67.3 & - & -\\
\hline 
word embedding & yes &  70.1  & 65.4 & \textbf{72.4} \\
word embedding & no & 72.3 & 65.4 & \textbf{73.0} \\
\hline
\end{tabular}
\caption{Ambiguity prediction accuracy for the supersense dataset.
Norm: L2-normalizing the vectors.}
\tablabel{supersense}

\end{table}

\end{document}